\documentclass[letterpaper]{article}
\usepackage{aaai2027}
\usepackage[hyphens]{url}
\usepackage{graphicx}
\usepackage{natbib}
\usepackage{caption}
\usepackage{algorithm}
\usepackage{amsmath}
\usepackage{booktabs}
\usepackage{amssymb}
\usepackage{multirow}
\usepackage{algpseudocode}
\usepackage{colortbl}
\usepackage{array}
\usepackage{tabularx}
\usepackage[most]{tcolorbox}

\definecolor{fasAccent}{HTML}{1F5FA8}
\definecolor{fasOurs}{HTML}{EAF2FF}
\definecolor{fasStrong}{HTML}{E8F6ED}
\definecolor{fasGain}{HTML}{168A45}
\newcommand{\papertablefont}{\fontsize{7.2pt}{8.0pt}\selectfont}

\newcommand{\tcite}[1]{{\scriptsize~\cite{#1}}}
\newcolumntype{L}{>{\raggedright\arraybackslash}X}
\newcolumntype{Y}{>{\centering\arraybackslash}X}

\nocopyright

\title{FAS-R1: A Unified Multi-Task MLLM for Reasoning Face Anti-Spoofing}
\author{
Hongyang Wang\textsuperscript{\rm 1,\rm 2}\equalcontrib,
Yichen Shi\textsuperscript{\rm 3}\equalcontrib,
Hongrui Li\textsuperscript{\rm 1}\equalcontrib,
Yiru Huo\textsuperscript{\rm 4},
Jun Feng\textsuperscript{\rm 1}\corresponding,
Zitong Yu\textsuperscript{\rm 5}\corresponding
}
\affiliations{
\textsuperscript{\rm 1}Shijiazhuang Tiedao University\quad
\textsuperscript{\rm 2}Fudan University\quad
\textsuperscript{\rm 3}Shanghai Jiao Tong University\\
\textsuperscript{\rm 4}Yanshan University\quad
\textsuperscript{\rm 5}Great Bay University
}

\begin{document}

\maketitle

\begin{abstract}
Face anti-spoofing (FAS) is increasingly expected to provide not only bona fide/spoof decisions, but also attack semantics and image-grounded evidence for human inspection. Existing discriminative FAS models remain largely label-centric, while recent MLLM-based methods offer structured outputs but still rely mainly on supervised fine-tuning, often producing template-like rationales and weak optimization for difficult attacks. We propose FAS-R1, a two-stage reasoning-oriented MLLM framework for unified FAS prediction, covering authenticity classification, attack-type recognition and spoof-region localization. FAS-R1 first uses FAS-R1-23K, a high-quality long-CoT dataset, for cold-start supervised fine-tuning, and then performs FAS-specific GRPO post-training. Degradation-Simulated Augmentation (DSA) encourages stable spoof-cue reasoning across visual-quality shifts, while Difficulty-Aware GRPO (DA-GRPO) mitigates easy-sample dominance that may leave difficult task--attack groups under-optimized, especially for subtle or ambiguous attacks such as makeup and mask attacks. The main 3B FAS-R1 model achieves 98.75\% authenticity accuracy, 93.33\% attack-type accuracy, and 96.30/94.73\% AP@40/AP@50 in-domain. It also outperforms the compared systems in cross-domain authenticity generalization and answer-and-rationale quality. Experiments with different base models further show favorable scaling behavior. The code will be released soon.
\end{abstract}

\section{Introduction}
\label{sec:intro}

Face anti-spoofing (FAS) protects face recognition systems against print, replay, 3D-mask, and partial presentation attacks~\cite{yu2022deep}. In payment, access control, and identity verification, FAS errors can directly affect security. Practical deployments further introduce heterogeneous cameras, illumination, compression, resolution, and sensor noise. A reliable FAS system should therefore go beyond bona fide/spoof prediction and expose attack semantics with supporting visual evidence.

\begin{figure}[t]
    \centering
    \includegraphics[width=1\linewidth]{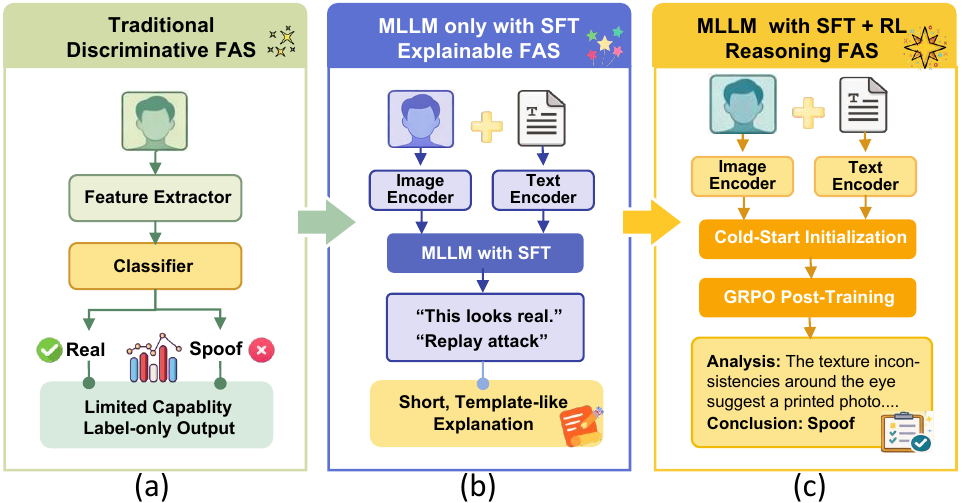}
    \caption{Comparison of three FAS paradigms: (a) discriminative models mainly output binary decisions; (b) SFT-based MLLMs provide structured but often template-style explanations; (c) FAS-R1 combines SFT and RL to produce structured predictions with inspectable rationales.}
    \vspace{-4mm}
    \label{fig:intro}
\end{figure}

Most traditional FAS methods are optimized for label-level discrimination. As illustrated in Fig.~\ref{fig:intro}(a), discriminative models can achieve strong binary classification performance~\cite{he2016resnet,Yu2020CDCN,wang2022patchnet}, but usually return only a class score. Such label-centric outputs provide limited attack semantics or spatial evidence, making failures difficult to inspect.

\begin{figure*}[t]
    \centering
    \includegraphics[width=1\textwidth]{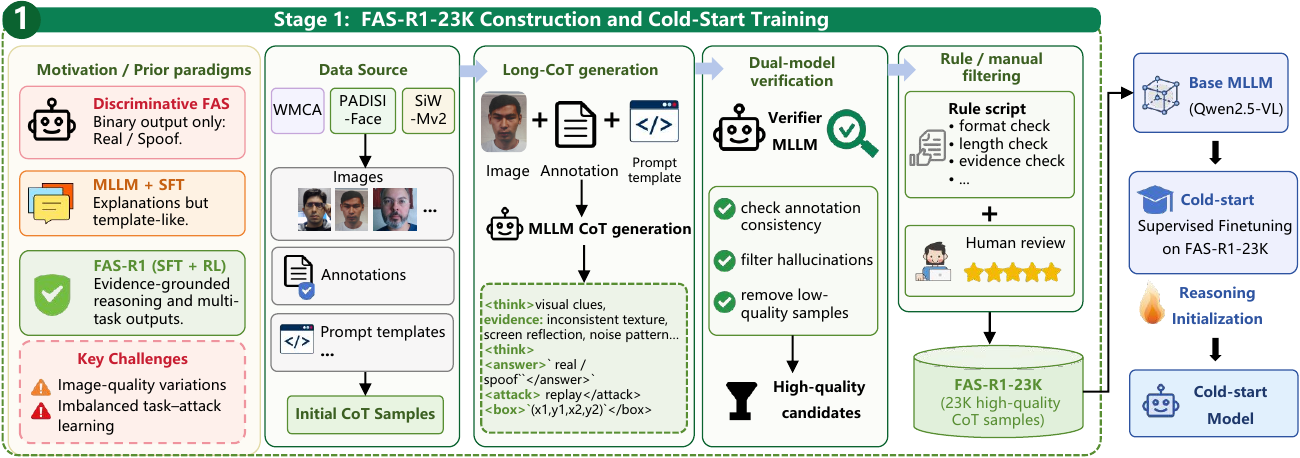}
    \caption{Stage 1: long-CoT cold start with annotation-constrained generation, external verification, and rule/manual filtering.}
    \label{fig:framework_overview_stage1}
    \vspace{-2mm}
\end{figure*}

Vision-language and MLLM-based FAS methods begin to address this gap by introducing textual semantics and natural-language rationales~\cite{shi2025shield,zhang2025interpretable}. FaceShield further studies a unified MLLM interface for authenticity classification, attack-type recognition, and attack-region localization~\cite{wang2025faceshield}. However, these systems are mainly driven by supervised fine-tuning (SFT). As illustrated in Fig.~\ref{fig:intro}(b), SFT teaches the model to follow an output format but can also encourage short, recurring explanations. The resulting rationales may look structured while remaining weakly grounded in image-specific evidence.

Recent FAS studies have also explored chain-of-thought supervision, tool-augmented reasoning, and reinforcement fine-tuning~\cite{zhang2026harnessingcotfas,jiang2025tasksolving,zhang2026tarfash,ma2026pafas}, showing the promise of post-training beyond SFT. Nevertheless, generic RL does not fully match FAS: models must learn stable spoof cues across visual-quality variations, and training can be dominated by easy samples. Easy cases may be quickly optimized, whereas complex task--attack cases receive weak corrective signals and remain mislearned, making hard-to-distinguish spoof patterns remain confused. Effective reasoning-oriented FAS therefore needs post-training aware of both cue stability and task--attack difficulty.

\begin{figure*}[t]
    \centering
    \includegraphics[width=1\textwidth]{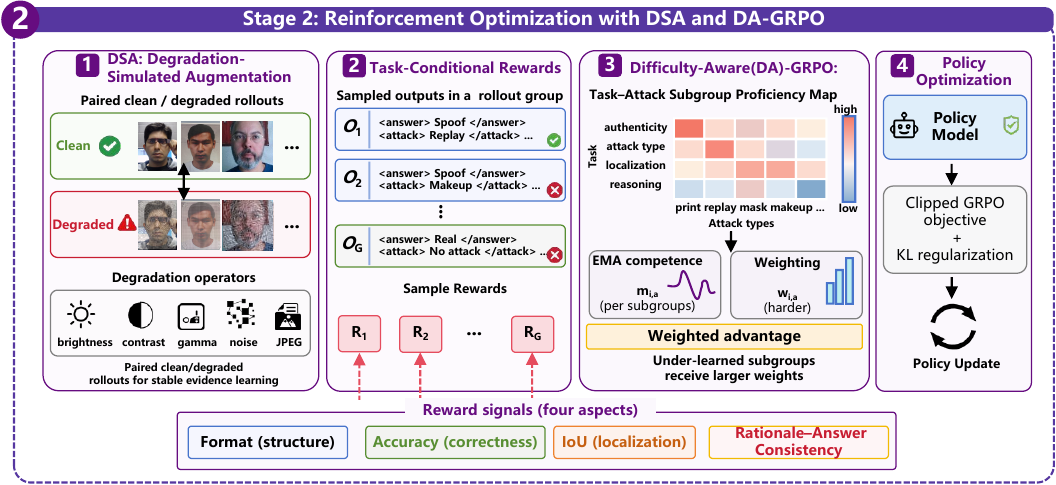}
    \caption{Stage 2: FAS-specific reinforcement optimization. DSA constructs paired clean/degraded rollouts, while DA-GRPO reweights task--attack subgroups according to online proficiency.}
    \label{fig:framework_overview_stage2}
    \vspace{-2mm}
\end{figure*}

Motivated by these observations, we propose FAS-R1, a two-stage reasoning-oriented MLLM framework for structured and inspectable FAS prediction. FAS-R1 uses a shared generation interface for authenticity classification, attack-type recognition, coarse spoof-region localization, and rationale generation, while improving evidence grounding within this interface. Stage 1 uses FAS-R1-23K for high-quality long-CoT cold-start SFT. Stage 2 performs FAS-specific GRPO: DSA groups clean/degraded rollouts to encourage stable spoof-cue reasoning under visual-quality shifts, and DA-GRPO redirects updates to persistently unreliable task--attack groups for complex samples. Fig.~\ref{fig:framework_overview_stage1} and Fig.~\ref{fig:framework_overview_stage2} summarize the two-stage pipeline. Our contributions are threefold:
\begin{itemize}
    \item We introduce FAS-R1, a two-stage reasoning-oriented MLLM framework for evidence-grounded structured FAS, together with FAS-R1-23K, a 22,996-sample high-quality CoT dataset.
    \item We develop FAS-specific GRPO with DSA for stable spoof-cue learning and DA-GRPO for easy-sample-dominated task--attack optimization.
    \item FAS-R1 achieves strong multi-task, cross-domain, localization, scaling, and answer-and-rationale results on WMCA\cite{george2019wmca}, PADISI-Face\cite{rostami2021padisi}, and SiW-Mv2\cite{guo2022siwmv2}.
\end{itemize}

\section{Related Work}
\label{sec:related_work}

\subsection{Face Anti-Spoofing}

\subsubsection{Traditional FAS}
Traditional FAS methods first rely on hand-crafted texture descriptors and shallow classifiers, such as micro-texture analysis and local binary patterns~\cite{Maatta2011MicroTexture,Chingovska2012LBP}. Deep models then learn spoof-related representations end-to-end~\cite{AtoumIJCB2017,GeorgeICB2019,YangCVPR2019}, and open-set or domain-generalization methods further improve transfer with domain-invariant representations, gradient alignment, and face-security pretraining~\cite{LiuCVPR2019DeepTree,Liu2023UDGFAS,le2024gacfas,wang2025fsfm}. These studies establish strong discriminative baselines, but most of them still treat FAS mainly as label prediction. As a result, they offer limited attack semantics or spatial evidence when the prediction is wrong or ambiguous. FAS-R1 keeps the discriminative goal of FAS, but extends the output space to attack semantics, coarse regions, and image-specific rationales in a unified generative interface.

\subsubsection{Vision-Language FAS}
Vision-language FAS methods add semantic supervision to this label-centric paradigm. CLIP-based approaches align facial observations with textual concepts or local/global visual-language correspondences for cross-domain recognition~\cite{Radford2021CLIP,Srivatsan2023FLIP,Liu2024CFPLFAS,Liu2024BUDoPT,Mu2023TeGDG,yu2025mvpfas}, but their outputs are still usually classification-oriented. MLLM-based studies further evaluate or generate natural-language FAS explanations~\cite{Liu2023LLaVA,shi2025shield,zhang2025interpretable,zhang2026harnessingcotfas}. FaceShield is an important step because it unifies authenticity classification, attack-type recognition, and attack-region localization in one MLLM framework~\cite{wang2025faceshield}, yet its SFT-centered training can still produce short, template-like rationales with limited optimization of sampled reasoning trajectories. Recent task-solving reinforcement fine-tuning, tool-augmented reasoning, and path-augmented RL methods explore post-training for FAS~\cite{jiang2025tasksolving,zhang2026tarfash,ma2026pafas}, but they do not explicitly target stable spoof-cue learning under visual-quality shifts or easy-sample-dominated optimization. FAS-R1 addresses these gaps with a high-quality long-CoT cold start and FAS-specific GRPO components for cue stability and hard subgroup learning.

\section{Method}
\label{sec:method}

\begin{table}[t]
\centering
\caption{MLLM-based FAS datasets. FAS-R1-23K uniquely combines authenticity, attack-type, localization, and long-CoT annotations.}
\label{tab:mllm_fas_data_compare}
\papertablefont
\setlength{\tabcolsep}{1.6pt}
\renewcommand{\arraystretch}{0.92}
\begin{tabularx}{\columnwidth}{@{}lccccc@{}}
\toprule
\textbf{Resource} & \textbf{Scale} & \textbf{Bona fide} & \textbf{Attack} & \textbf{Region} & \textbf{Long} \\
& & \textbf{/spoof} & \textbf{type} & \textbf{loc.} & \textbf{CoT} \\
\midrule
I-FAS~\cite{zhang2025interpretable} & 12 ds. & \checkmark & -- & -- & -- \\
FaceCoT~\cite{zhang2026harnessingcotfas} & 1.08M & \checkmark & -- & -- & \checkmark \\
PA-FAS~\cite{ma2026pafas} & 800 paths & \checkmark & -- & -- & Path \\
FaceShield~\cite{wang2025faceshield} & 45K & \checkmark & \checkmark & \checkmark & -- \\
\rowcolor{fasStrong}
\textbf{FAS-R1-23K} & \textbf{23K} & \textbf{\checkmark} & \textbf{\checkmark} & \textbf{\checkmark} & \textbf{\checkmark} \\
\bottomrule
\end{tabularx}
\end{table}

\begin{algorithm}[t]
\small
\caption{Two-stage FAS-R1 training}
\label{alg:fasr1}
\begin{algorithmic}[1]
\Require Policy $\pi_\theta$, reference $\pi_{\mathrm{ref}}$, CoT set $\mathcal{D}_{\mathrm{cot}}$, RL set $\mathcal{D}$, degradation operator $A$, group size $G$
\State Train $\pi_\theta$ on $\mathcal{D}_{\mathrm{cot}}$ by SFT
\For{each RL step}
    \State Sample $\{(p_i,I_i,t_i,a_i)\}$ from $\mathcal{D}$
    \State Generate clean and degraded rollouts from $I_i$ and $A(I_i)$
    \State Compute task-specific rewards and GRPO advantages $\hat A_{i,j}$
    \State Update the EMA proficiency of each observed $(t_i,a_i)$ subgroup
    \State Set $\widetilde A_{i,j}=w_{t_i,a_i}\hat A_{i,j}$ after warm-up
    \State Update $\pi_\theta$ with the clipped GRPO objective
\EndFor
\end{algorithmic}
\end{algorithm}

\begin{figure*}[t]
    \centering
    \includegraphics[width=0.82\textwidth]{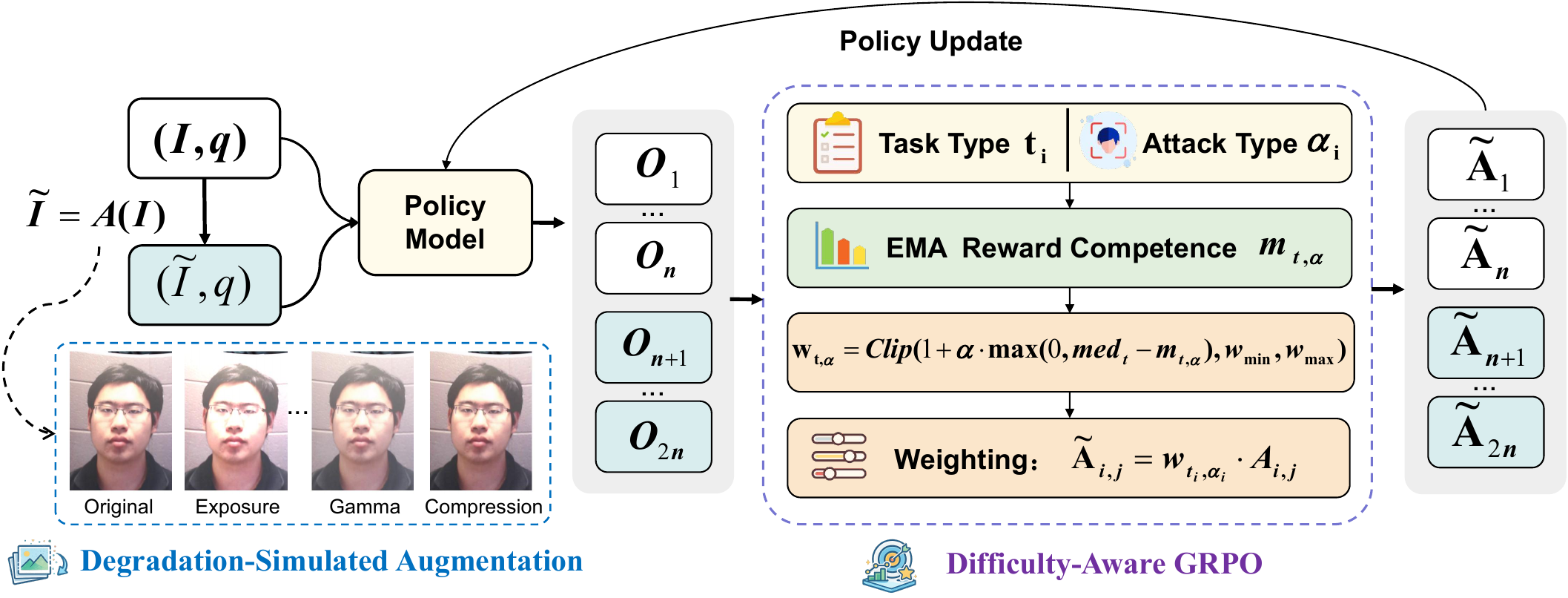}
    \caption{FAS-specific GRPO optimization. DSA mixes clean and degraded trajectories in each rollout group, and DA-GRPO estimates task--attack proficiency to rescale advantages after warm-up.}
    \label{fig:da-grpo}
    \vspace{-4mm}
\end{figure*}

\subsection{Stage 1: Cold Start}
\label{sec:cold_start}

\subsubsection{Data Construction}

As summarized in Fig.~\ref{fig:framework_overview_stage1}, FAS-R1-23K is built from WMCA~\cite{george2019wmca}, PADISI-Face~\cite{rostami2021padisi}, and SiW-Mv2~\cite{guo2022siwmv2}. Gemini 2.5 Flash~\cite{gemini25flash} generates annotation-constrained candidate rationales, and GPT-5~\cite{gpt5} verifies answer correctness and annotation--rationale consistency. Rule-based filters check format, answer type, prompt leakage, contradictions, and irrelevant reasoning, followed by manual inspection of flagged samples. The final corpus retains 22,996 samples in the \texttt{<think>...</think><answer>...</answer>} format.

Authenticity labels, attack categories, and localization annotations are inherited from the original datasets rather than generated by MLLMs. The rationales must contain task-specific visual evidence, and localization uses the manually annotated attack-region boxes; global attacks use the visible face or presentation region as the target. FAS-R1-23K is an annotation-constrained rationale-supervision corpus rather than a human-authored explanation benchmark.

This design targets a supervision gap in current MLLM-based FAS data. As shown in Table~\ref{tab:mllm_fas_data_compare}, FAS-R1-23K uniquely combines authenticity, explicit attack-type QA, localization, and long-CoT supervision with annotation verification.

\subsubsection{Model Training}
We perform full-parameter SFT of Qwen2.5-VL-3B and Qwen2.5-VL-7B~\cite{bai2025qwen25vl} with \texttt{LLaMA-Factory}~\cite{zheng2024llamafactory}. The vision encoder, multimodal projector, and language model are all trainable. Unless specified otherwise, SFT uses a learning rate of $1.0\times10^{-6}$ for two epochs. This stage establishes task semantics, output structure, and domain-specific rationale generation before on-policy optimization.




\subsection{Stage 2: Reinforcement Learning}
\label{sec:rl_stage}

After cold-start SFT, we apply GRPO-based optimization. Prior FAS reasoning models have explored reinforcement fine-tuning or path/tool-augmented reasoning~\cite{jiang2025tasksolving,zhang2026tarfash,ma2026pafas}, but generic GRPO does not explicitly handle two FAS-specific issues: spoof cues should remain stable under visual-quality shifts, and easy subgroups can dominate optimization while harder task--attack cases receive weak corrective signals. We therefore introduce Degradation-Simulated Augmentation (DSA) for paired clean/degraded rollouts and Difficulty-Aware GRPO (DA-GRPO) for adaptive task--attack optimization.

Algorithm~\ref{alg:fasr1} summarizes the training procedure, and Fig.~\ref{fig:da-grpo} illustrates their interaction.

\subsubsection{Reward Design}

We use task-conditional rewards for output format, answer correctness, localization IoU, and GPT-5-based rationale--answer consistency. All RL variants share the same verifier, reward definitions, prompts, and weights; only rollout construction and advantage weighting differ. The normalized rewards are

\begin{equation}
R_{\mathrm{QA}}=
\tilde\omega_f r_{\mathrm{fmt}}^{\mathrm{QA}}
+\tilde\omega_a r_{\mathrm{acc}}
+\tilde\omega_t r_{\mathrm{think}},
\end{equation}

\begin{equation}
R_{\mathrm{Loc}}=
\tilde\omega_b r_{\mathrm{fmt}}^{\mathrm{box}}
+\tilde\omega_i r_{\mathrm{IoU}}.
\end{equation}

We use QA weights $0.1/0.5/0.4$, localization weights $0.2/0.8$, and $\tau=0.5$.

\begin{table*}[tbp]
\centering
\caption{In-domain comparison on authenticity classification, attack-type recognition, and attack-region localization (\%). AP@$t$ denotes single-box IoU-threshold success rate. Best and second-best results are bold and underlined.}
\label{tab:unified_three_tasks}

\papertablefont
\setlength{\tabcolsep}{2.6pt}
\renewcommand{\arraystretch}{0.92}

\begin{tabularx}{\textwidth}{@{}ll*{5}{Y}@{}}
\toprule
\multirow{2}{*}{Category} & \multirow{2}{*}{Model} &
\multicolumn{2}{c}{\textbf{Coarse-grained}} &
\multicolumn{1}{c}{\textbf{Fine-grained}} &
\multicolumn{2}{c}{\textbf{Localization}} \\
\cmidrule(lr){3-4}\cmidrule(lr){5-5}\cmidrule(lr){6-7}
& & ACC$\uparrow$ & HTER$\downarrow$ & ACC$\uparrow$ & AP@40$\uparrow$ & AP@50$\uparrow$ \\
\midrule

\multirow{3}{*}{Trad.}
& ResNet\tcite{he2016resnet}                     & 97.55 & 2.32 & --    & --    & -- \\
& PatchNet\tcite{wang2022patchnet}               & 98.22 & 1.78 & --    & --    & -- \\
& CoOp\tcite{zhou2022coop}                       & \underline{98.73} & \underline{1.27} & -- & -- & -- \\
\midrule

\multirow{10}{*}{MLLM}
& LLaVA\tcite{Liu2023LLaVA}                      & 65.54 & 27.76 & 16.39 & --    & -- \\
& Qwen-VL\tcite{bai2023qwen-vl}                  & 51.94 & 38.70 & 16.55 &  2.07 &  1.49 \\
& MiniGPT-4\tcite{zhu2023minigpt4}               & 26.86 & 65.50 & 19.51 & --    & -- \\
& Lenna\tcite{wei2023lenna}                      & --    & --    & --    & 37.77 & 35.41 \\
& Sphinx\tcite{lin2023sphinx}                    & --    & --    & --    & 47.86 & 46.30 \\
& Bunny\tcite{he2024bunny}                       & 81.20 & 17.87 & 27.03 & 73.50 & 71.65 \\
& Claude-Sonnet-4.5\tcite{anthropic2025claude45} & 71.33 & 28.67 & 58.33 & 55.68 & 51.13 \\
& GPT-5.2\tcite{openai2025gpt52systemcard}       & 69.33 & 30.67 & 44.00 & 48.10 & 43.04 \\
& Gemini-3-Pro\tcite{google2025gemini3pro}       & 93.27 & 6.80  & 79.67 & 89.66 & 89.66 \\
& FaceShield\tcite{wang2025faceshield}           & 95.95 & 3.61  & \underline{93.24} & 73.79 & 70.23 \\
& PA-FAS\tcite{ma2026pafas}                      & 97.93    & 2.07    & 91.82    & \underline{92.62}    & \underline{91.30} \\
\midrule
\rowcolor{fasStrong}
Ours & \textbf{FAS-R1 (Ours)}                    & \textbf{{98.75} }& \textbf{1.17} & \textbf{93.33} & \textbf{96.30} & \textbf{94.73} \\
\bottomrule
\end{tabularx}
\vspace{-2mm}

\end{table*}

\subsubsection{Degradation-Simulated Augmentation}
\label{sec:dsa}

Spoof cues should be learned from stable evidence rather than incidental image quality. However, applying degradation to all rollouts may remove clean visual references and destabilize GRPO optimization. DSA addresses this issue by placing paired clean and synthetically degraded trajectories within the same rollout group. This makes the policy compare rewards across clean and perturbed views of the same annotated sample, encouraging the reasoning trajectory to retain attack-relevant cues instead of overfitting to a specific image quality.

For each image--prompt pair $(I_i,p_i)$, clean trajectories are sampled as

\begin{equation}
\{y^c_{i,k}\}_{k=1}^{n_c}
\sim
\pi_\theta(\cdot|p_i,I_i),
\end{equation}

and degraded trajectories are sampled as

\begin{equation}
\{y^d_{i,k}\}_{k=1}^{n_d}
\sim
\pi_\theta(\cdot|p_i,A(I_i)),
\end{equation}

where $A$ applies moderate brightness, contrast, gamma, Gaussian noise, and JPEG perturbations. Clean and degraded trajectories share identical annotations and reward functions, and are combined into the same rollout group. DSA does not introduce a new image-degradation model; its distinction lies in placing clean and synthetically degraded views of the same annotated input within the same on-policy rollout group.

Let $\mathcal G_i$ denote the union of clean and degraded trajectories for sample $i$, with $G=n_c+n_d$. We optimize this mixed rollout group with a clipped group-relative objective~\cite{shao2024deepseekmath}:

\begin{equation}
\begin{aligned}
\mathcal J_{\mathrm{DSA}}
=
\mathbb E_{(p_i,I_i)}
\left[
\frac{1}{G}\sum_{j=1}^{G}
\min(r_{i,j}\hat A_{i,j},\bar r_{i,j}\hat A_{i,j})
\right. \\
\left.
-\beta
D_{\mathrm{KL}}(\pi_\theta\|\pi_{\mathrm{ref}})
\right],
\end{aligned}
\end{equation}

where $\mathcal G_i=\{y^c_{i,k}\}_{k=1}^{n_c}\cup\{y^d_{i,k}\}_{k=1}^{n_d}$, $r_{i,j}$ is the policy ratio with token indices suppressed, $\bar r_{i,j}=\mathrm{clip}(r_{i,j},1-\epsilon,1+\epsilon)$, and $\hat A_{i,j}=(R_{i,j}-\mu_i)/(\sigma_i+\delta)$ is normalized within $\mathcal G_i$. Different from offline augmentation, DSA directly modifies on-policy exploration while maintaining clean visual anchors and encouraging consistent task behavior across clean and perturbed views.

\subsubsection{Difficulty-Aware GRPO}
\label{sec:dagrpo}

Multi-task FAS exhibits heterogeneous learning dynamics: easy subgroups can be optimized quickly, while difficult task--attack cases may keep receiving weak or misleading corrective signals under vanilla GRPO. DA-GRPO addresses this imbalance by defining difficulty at the task--attack subgroup level. Instead of treating every low-reward sample equally, it tracks whether a semantic subgroup remains unreliable relative to other attacks within the same task and then increases its on-policy learning signal.

Let $\bar R_{t,a}^{(s)}$ denote the mean reward of subgroup $(t,a)$ at training step $s$. The online proficiency is updated as

\begin{equation}
m_{t,a}^{(s)}
=
\rho m_{t,a}^{(s-1)}
+
(1-\rho)\bar R_{t,a}^{(s)}.
\end{equation}

Difficulty is compared only within the same task to avoid mixing reward scales. Given the median proficiency $\mathrm{med}_t$ across attack categories in task $t$ as a robust within-task reference, the preliminary weight is

\begin{equation}
w'_{t,a}
=
\mathrm{Clip}
(
1+\lambda\max(0,\mathrm{med}_t-m_{t,a}),
w_{\min},w_{\max}
).
\end{equation}

Weights are normalized to unit mean within each task. After warm-up, the GRPO advantage becomes

\begin{equation}
\widetilde A_{i,j}
=
\frac{w'_{t_i,a_i}}
{\mathbb E_{a|t_i}[w'_{t_i,a}]}
\hat A_{i,j}.
\end{equation}

The default settings use a five-step warm-up, $\rho=0.98$, $\lambda=2.0$, $w_{\min}=0.7$, and $w_{\max}=2.0$. We update a subgroup only when at least four prompts are observed; clipping, warm-up, and unit-mean normalization stabilize the policy-update scale.

Unlike conventional hard-example weighting, DA-GRPO estimates difficulty over semantic task--attack subgroups, compares proficiency within each task, and reweights on-policy GRPO advantages rather than supervised losses.

The two stages therefore set up the experimental questions: whether FAS-R1 can improve multi-task accuracy, preserve cross-domain generalization, produce inspectable rationales, and isolate the effects of DSA and DA-GRPO.

\begin{table*}[tbp]
\centering
\caption{Cross-domain authenticity generalization (\%). W, P, and S denote WMCA, PADISI-Face, and SiW-Mv2; each protocol trains on two datasets and tests on the held-out dataset.}
\label{tab:three_protocols_fit}
\papertablefont
\setlength{\tabcolsep}{2.6pt}
\renewcommand{\arraystretch}{0.92}

\begin{tabularx}{\textwidth}{@{}l*{6}{Y}@{}}
\toprule
& \multicolumn{2}{c}{\textbf{W \& S$\rightarrow$P}} &
  \multicolumn{2}{c}{\textbf{W \& P$\rightarrow$S}} &
  \multicolumn{2}{c}{\textbf{S \& P$\rightarrow$W}} \\
\cmidrule(lr){2-3}\cmidrule(lr){4-5}\cmidrule(lr){6-7}
\textbf{Methods} &
\textbf{ACC(\%)$\uparrow$} & \textbf{HTER(\%)$\downarrow$} &
\textbf{ACC(\%)$\uparrow$} & \textbf{HTER(\%)$\downarrow$} &
\textbf{ACC(\%)$\uparrow$} & \textbf{HTER(\%)$\downarrow$} \\
\midrule
ResNet\tcite{he2016resnet}           & 46.12 & 50.00 & 53.36 & 49.16 & 74.01 & 29.75 \\
PatchNet\tcite{wang2022patchnet}     & 77.18 & 22.87 & 56.16 & 45.37 & 78.15 & 41.50 \\
IADG\tcite{zhou2023iadg}             & 72.96 & 27.01 & 57.20 & 42.81 & 78.55 & 26.27 \\
FAS-AUG\tcite{cai2024fasaug}         & 91.70 & 7.30  & 88.20 & 11.70 & 87.90 & 13.10 \\
FaceShield\tcite{wang2025faceshield} & 88.40 & 12.14 & 92.63 & 7.58 & 91.91 & \underline{5.80} \\
PA-FAS\tcite{ma2026pafas}            & \textbf{92.74}    & \textbf{6.14}    & \underline{92.68}    & \underline{6.72}    & \underline{92.83}    & 6.01 \\
\rowcolor{fasStrong}
\textbf{FAS-R1 (Ours)}                 & \underline{92.39} & \underline{6.35} & \textbf{93.42} & \textbf{5.52} & \textbf{93.49} & \textbf{5.34} \\
\bottomrule
\end{tabularx}
\end{table*}

\section{Experiments}
\label{sec:experiment}

We test multi-task accuracy, cross-domain transfer, rationale quality, and the effects of DSA and DA-GRPO. Unless otherwise specified, the main tables report Qwen2.5-VL-3B, while Table~\ref{tab:model_scale} additionally reports 7B.



\subsection{Evaluation Protocols}

Following FaceShield~\cite{wang2025faceshield}, we evaluate three tasks on WMCA (W)~\cite{george2019wmca}, PADISI-Face (P)~\cite{rostami2021padisi}, and SiW-Mv2 (S)~\cite{guo2022siwmv2}. For in-domain evaluation, the merged images are split at the image level into training, validation, and test sets with an 8:1:1 ratio. FAS-R1-23K is constructed only from the training set, while validation and test images are excluded from corpus construction. For cross-domain evaluation, both corpus construction and training use source domains only. We report ACC/HTER, ACC, and AP@40/AP@50 for the three tasks, respectively. Trainable MLLM baselines are evaluated using the same training data and prompt templates. As other MLLM-based FAS methods have not released their source code or model checkpoints, the task-specific evaluation is primarily conducted against FaceShield and PA-FAS.

\subsection{Implementation Details}

The Qwen2.5-VL-3B/7B backbones~\cite{bai2025qwen25vl} are trained on four NVIDIA A800 80GB GPUs with two full-parameter epochs, six rollouts per sample, learning rate $1\times10^{-6}$, and global batch size 128. DSA is applied to half of the rollouts, and all RL variants share reward functions and weights. Generic MLLMs are prompt-only references, while FaceShield~\cite{wang2025faceshield} and PA-FAS~\cite{ma2026pafas} are trained with FAS-R1-23K.

\subsection{In-Domain Evaluation}
Table~\ref{tab:unified_three_tasks} evaluates whether FAS-R1 preserves classification accuracy while adding attack semantics and spatial evidence.

\noindent\textbf{Coarse-grained classification.} The 3B FAS-R1 reaches 98.75\% ACC and 1.17\% HTER, matching strong discriminative baselines while also producing attack semantics and rationales.

\noindent\textbf{Fine-grained classification.} FAS-R1 reaches 93.33\% ACC, slightly above FaceShield~\cite{wang2025faceshield}. Bona fide samples are correct by construction, so this score should be read together with authenticity accuracy.

\noindent\textbf{Attack-region localization.} FAS-R1 improves over FaceShield from 73.79/70.23 to 96.30/94.73 AP@40/AP@50 and outperforms PA-FAS at 92.62/91.30.

\subsection{Cross-Domain Generalization}
We next test transfer to unseen acquisition conditions. In Table~\ref{tab:three_protocols_fit}, the 3B FAS-R1 achieves competitive results across all protocols; together with Table~\ref{tab:model_scale}, the strongest FAS-R1 checkpoint obtains the best compared result on all three settings. We focus on authenticity because attack taxonomies and spatial annotations are not fully aligned across datasets.


\subsection{Answer-and-Rationale Quality}
Accuracy alone does not indicate whether the generated rationales support the final decision. Following VERITAS~\cite{tan2026veritas}, Claude-Sonnet-4.5~\cite{anthropic2025claude45} and Gemini-3-Pro~\cite{google2025gemini3pro} assess answer correctness, visual relevance, coherence, and clarity, with pairwise preferences converted into Elo ratings. Both judge models are also included as candidate systems and assign FAS-R1 higher scores than their own outputs, further supporting consistency across judges. FAS-R1 achieves the highest scores (4.79/4.70) and Elo rating (1803.34).

\begin{figure}[tbp]
\centering
\includegraphics[width=1\linewidth]{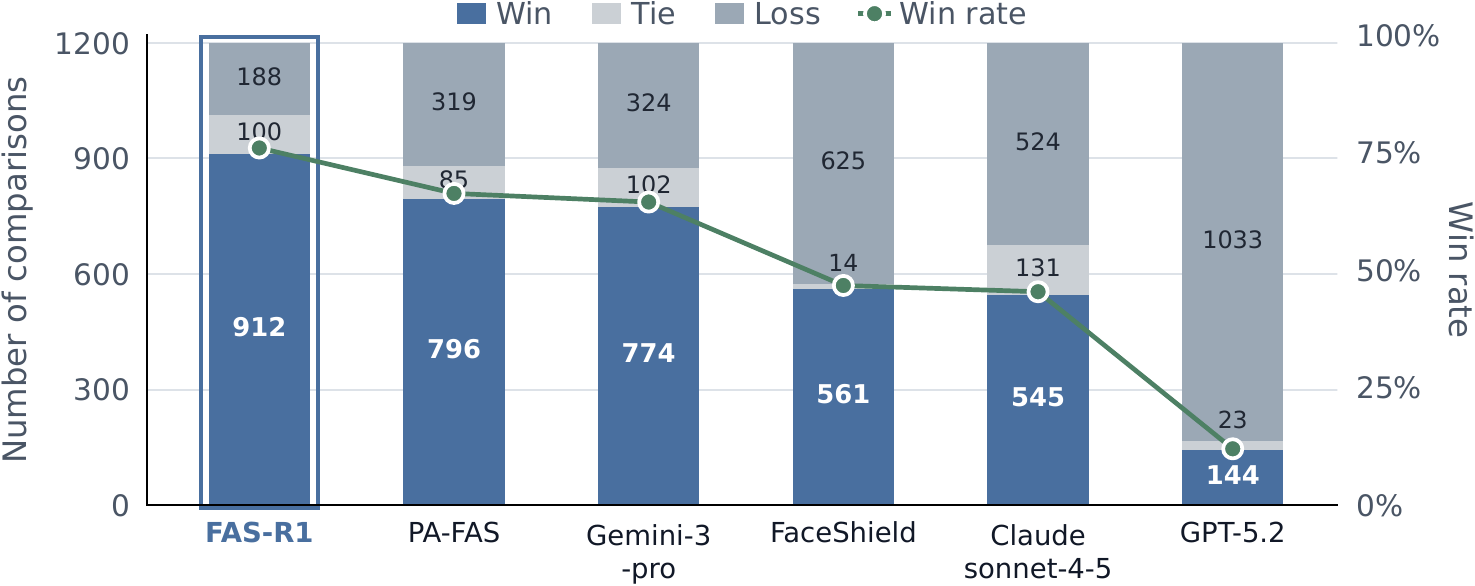}
\caption{Automated pairwise answer-and-rationale comparison. Bars show wins, ties, and losses over 1,200 comparisons, and the line reports win rate.}
\vspace{-5mm}
\label{fig:elo_visualization}
\end{figure}

\begin{table}[!htbp]
\centering
\caption{Answer-and-rationale quality. Judge scores use a 1--5 scale; pairwise preferences are reported as Elo ratings.}
\label{tab:inference_quality_elo}
\papertablefont
\setlength{\tabcolsep}{2.6pt}
\renewcommand{\arraystretch}{0.92}
\begin{tabularx}{\columnwidth}{@{}l*{3}{Y}@{}}
\toprule
\multirow{2}{*}{Model} & \multicolumn{2}{c}{Judge scores} & \multirow{2}{*}{Elo(Initial 1500)} \\
\cmidrule(lr){2-3}
& Claude-Sonnet-4.5 & Gemini-3-Pro & \\
\midrule
GPT-5.2            & 2.23 & 2.16 & 1203.26 \\
Claude-Sonnet-4.5  & 3.40 & 3.48 & 1298.42 \\
Gemini-3-Pro       & \underline{4.26} & \underline{4.31} & \underline{1687.34} \\
FaceShield         & 3.56 & 3.67 & 1385.25 \\
PA-FAS\tcite{ma2026pafas} & 4.35 & 4.41 & 1704.82 \\
\rowcolor{fasStrong}
\textbf{FAS-R1 (Ours)} & \textbf{4.79} & \textbf{4.70} & \textbf{1803.34} \\
\bottomrule
\end{tabularx}
\end{table}

\subsection{Ablation Study}

\begin{figure}[tbp]
    \centering
    \includegraphics[width=\linewidth]{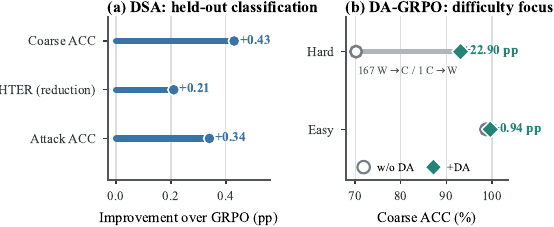}
    \caption{FAS-specific RL diagnostics: (a) DSA improves held-out metrics over GRPO; (b) DA-GRPO substantially improves performance on challenging samples.}
    \label{fig:training_reward_gain_3b7b}
\end{figure}

Table~\ref{tab:ablation} summarizes the progressive component ablation,
Fig.~\ref{fig:training_reward_gain_3b7b} provides further analysis of DSA and DA-GRPO,
and Table~\ref{tab:grpo_variant_ablation} compares alternative GRPO variants.
We analyze the contribution of each component below.

\paragraph{Two-Stage Training Strategy.}
The progression from the base model to cold-start SFT and GRPO shows that SFT establishes the structured task interface, while on-policy optimization further improves the performance of three tasks.

\begin{table}[tbp]
\centering
\caption{Component ablation of FAS-R1 with the 3B backbone. Best and second-best results are bold and underlined.}
\label{tab:ablation}
\fontsize{6.4pt}{7.1pt}\selectfont
\setlength{\tabcolsep}{0.7pt}
\renewcommand{\arraystretch}{0.90}

\begin{tabularx}{\columnwidth}{
@{}
>{\raggedright\arraybackslash}p{0.14\columnwidth}
*{4}{>{\centering\arraybackslash}p{0.052\columnwidth}}
*{5}{Y}
@{}}
\toprule
\multirow{2}{*}{\textbf{Variant}} &
\multicolumn{4}{c}{\textbf{Components}} &
\multicolumn{2}{c}{\textbf{Coarse}} &
\textbf{Fine} &
\multicolumn{2}{c}{\textbf{Localization}} \\
\cmidrule(lr){2-5}
\cmidrule(lr){6-7}
\cmidrule(lr){8-8}
\cmidrule(lr){9-10}
& \textbf{SFT}
& \textbf{RL}
& \textbf{DSA}
& \textbf{DA}
& \textbf{ACC$\uparrow$}
& \textbf{HTER$\downarrow$}
& \textbf{ACC$\uparrow$}
& \textbf{AP@40$\uparrow$}
& \textbf{AP@50$\uparrow$} \\
\midrule

Base
& -- & -- & -- & --
& 69.72 & 28.75 & 54.88 & 11.39 & 8.43 \\

Cold-start
& \checkmark & -- & -- & --
& 93.82 & 4.07 & 90.71 & 93.79 & 92.11 \\

GRPO
& \checkmark & \checkmark & -- & --
& 94.02 & 3.98 & 92.57 & 95.58 & 94.29 \\

\rowcolor{fasOurs}
\textcolor{fasAccent}{\textbf{+DSA}}
& \checkmark & \checkmark & \checkmark & --
& \underline{94.45}
& \underline{3.77}
& \underline{92.91}
& \textbf{96.44}
& \underline{94.59} \\

\rowcolor{fasStrong}
\textbf{FAS-R1}
& \checkmark & \checkmark & \checkmark & \checkmark
& \textbf{98.75}
& \textbf{1.17}
& \textbf{93.33}
& \underline{96.30}
& \textbf{94.73} \\

\bottomrule
\end{tabularx}
\end{table}

\paragraph{Effect of DSA.}
Relative to GRPO, introducing DSA consistently improves all five metrics, with AP@40 increasing from 95.58\% to 96.44\%. This result supports the effectiveness of placing paired clean and degraded views within the same rollout group.

\paragraph{Effect of DA-GRPO.}
Adding DA-GRPO on top of GRPO+DSA increases coarse-grained ACC from 94.45\% to 98.75\% and reduces HTER from 3.77\% to 1.17\%. Fine-grained ACC also improves, while localization remains stable. The challenging-sample analysis in Fig.~\ref{fig:training_reward_gain_3b7b} further shows substantial performance gains.

\begin{figure*}[t]
    \centering
    \includegraphics[width=0.9\linewidth]{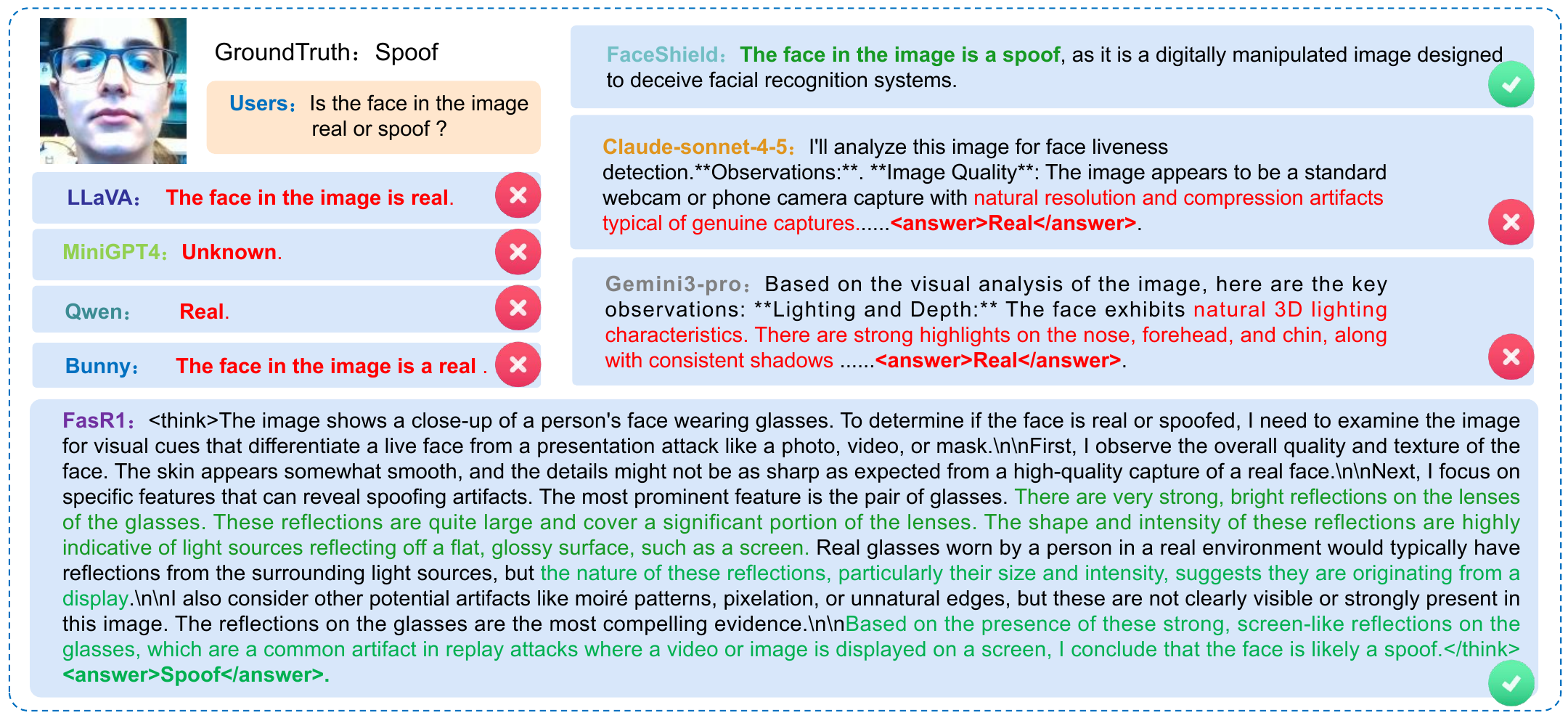}
    \caption{Authenticity example on a replay attack. FAS-R1 identifies image-specific reflective cues and correctly predicts spoof, while several baselines either misclassify the sample or provide generic descriptions.}
    \label{fig:qualitative_authenticity}
\end{figure*}

\begin{figure}[t]
    \centering
    \includegraphics[width=0.8\linewidth]{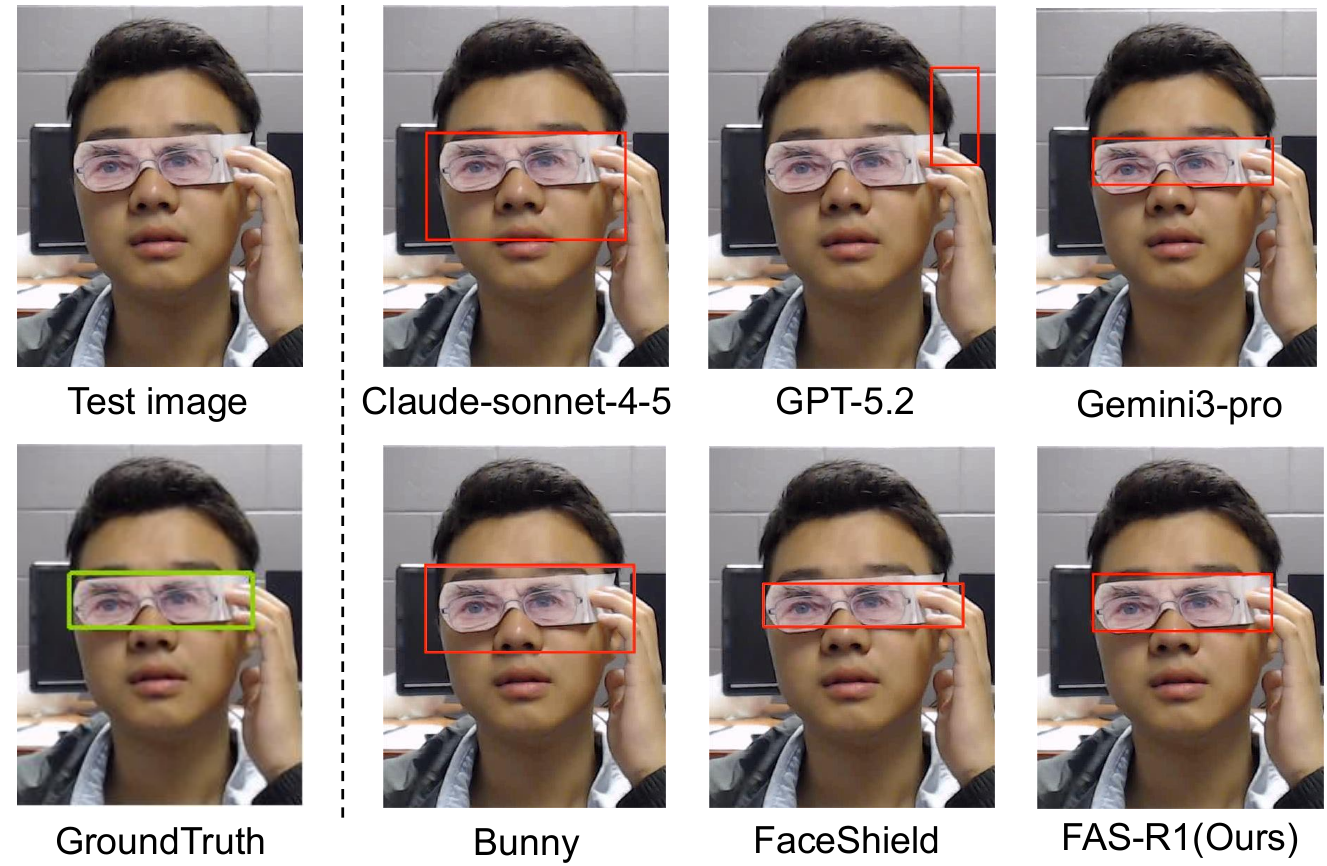}
    \caption{Localization example on a partial-eye attack.}
    \vspace{-2mm}
    \label{fig:qualitative_localization}
\end{figure}

\paragraph{Comparison with GRPO Variants.}
Under the same cold-start checkpoint and reward setting, neither DAPO\cite{yu2025dapo} nor GSPO\cite{zheng2025gspo} matches the overall performance of FAS-R1. This indicates that the gains are not obtained by simply replacing GRPO with a generic variant, but arise from the FAS-specific combination of DSA and DA-GRPO.

\begin{table}[!htbp]
\centering
\caption{Comparison with GRPO variants under the same setting. G+D denotes GRPO+DSA.}
\label{tab:grpo_variant_ablation}
\fontsize{6.8pt}{7.4pt}\selectfont
\setlength{\tabcolsep}{1.2pt}
\renewcommand{\arraystretch}{0.90}

\begin{tabularx}{\columnwidth}{@{}l*{5}{Y}@{}}
\toprule
\multirow{2}{*}{\textbf{Variant}} &
\multicolumn{2}{c}{\textbf{Coarse}} &
\textbf{Fine} &
\multicolumn{2}{c}{\textbf{Localization}} \\
\cmidrule(lr){2-3}
\cmidrule(lr){4-4}
\cmidrule(lr){5-6}
& \textbf{ACC$\uparrow$}
& \textbf{HTER$\downarrow$}
& \textbf{ACC$\uparrow$}
& \textbf{AP@40$\uparrow$}
& \textbf{AP@50$\uparrow$} \\
\midrule

GRPO
& 94.02 & 3.98 & 92.57 & 95.58 & 94.29 \\

DAPO
& 93.45 & \underline{3.56} & 91.98 & 96.09 & 94.52 \\

GSPO
& 92.27 & 4.43 & 91.23 & 95.22 & 93.78 \\

G+D
& \underline{94.45}
& 3.77
& \underline{92.91}
& \textbf{96.44}
& \underline{94.59} \\

\rowcolor{fasStrong}
\textbf{FAS-R1}
& \textbf{98.75}
& \textbf{1.17}
& \textbf{93.33}
& \underline{96.30}
& \textbf{94.73} \\

\bottomrule
\end{tabularx}
\end{table}


\paragraph{Model Scaling.} Table~\ref{tab:model_scale} shows that FAS-R1 performs strongly with the 3B backbone, while applying the strategy to 7B further improves all in-domain metrics and most cross-domain results, validating its effectiveness and scalability.

\begin{table}[t]
\centering
\caption{Backbone scaling on in-domain multi-task performance and cross-domain authenticity generalization.}
\label{tab:model_scale}
\papertablefont
\renewcommand{\arraystretch}{0.95}

\begin{tabularx}{\columnwidth}{@{}l*{5}{>{\centering\arraybackslash}X}@{}}
\toprule
\multicolumn{6}{c}{\textbf{In-domain}} \\
\midrule
\multirow{2}{*}{\textbf{Backbone}}
&
\multicolumn{2}{c}{\textbf{Coarse}}
&
\textbf{Fine}
&
\multicolumn{2}{c}{\textbf{Localization}}
\\
\cmidrule(lr){2-3}
\cmidrule(lr){4-4}
\cmidrule(lr){5-6}
&
ACC$\uparrow$
&
HTER$\downarrow$
&
ACC$\uparrow$
&
AP@40$\uparrow$
&
AP@50$\uparrow$
\\
\midrule

\rowcolor{fasOurs}
3B
& \underline{98.75}
& \underline{1.17}
& \underline{93.33}
& \underline{96.30}
& \underline{94.73} \\

\rowcolor{fasStrong}
7B
& \textbf{99.55}
& \textbf{0.32}
& \textbf{94.68}
& \textbf{97.07}
& \textbf{95.79} \\

\bottomrule
\end{tabularx}

\vspace{3mm}

\begin{tabularx}{\columnwidth}{@{}l*{6}{>{\centering\arraybackslash}X}@{}}
\toprule
\multicolumn{7}{c}{\textbf{Cross-domain (Coarse)}} \\
\midrule
\multirow{2}{*}{\textbf{Backbone}}
&
\multicolumn{2}{c}{\textbf{W\&S$\rightarrow$P}}
&
\multicolumn{2}{c}{\textbf{W\&P$\rightarrow$S}}
&
\multicolumn{2}{c}{\textbf{S\&P$\rightarrow$W}}
\\
\cmidrule(lr){2-3}
\cmidrule(lr){4-5}
\cmidrule(lr){6-7}
&
ACC$\uparrow$
&
HTER$\downarrow$
&
ACC$\uparrow$
&
HTER$\downarrow$
&
ACC$\uparrow$
&
HTER$\downarrow$
\\
\midrule

\rowcolor{fasOurs}
3B
& \underline{92.39}
& \underline{6.35}
& \underline{93.42}
& \underline{5.52}
& \textbf{93.49}
& \textbf{5.34} \\

\rowcolor{fasStrong}
7B
& \textbf{93.16}
& \textbf{6.04}
& \textbf{94.04}
& \textbf{5.08}
& \underline{93.26}
& \underline{5.65} \\

\bottomrule
\vspace{-4mm}
\end{tabularx}
\end{table}

\section{Qualitative Analysis}

\label{sec:visualization}

To further examine FAS-R1, we visualize its predictions for authenticity classification and attack-region localization. As shown in Fig.~\ref{fig:qualitative_authenticity}, FAS-R1 identifies localized reflective artifacts in a print attack and correctly predicts spoof, whereas several baselines either misclassify the sample as bona fide or provide generic descriptions weakly related to the image. In Fig.~\ref{fig:qualitative_localization}, FAS-R1 localizes the annotated eye region of a partial-eye attack, while other predictions are broader or spatially shifted. Together, these examples demonstrate the strong qualitative performance of FAS-R1.

\section{Conclusion}

In this work, we presented FAS-R1, a two-stage reasoning-oriented MLLM framework that advances face anti-spoofing from label-centric classification to evidence-grounded reasoning. First, we constructed FAS-R1-23K, a high-quality long-CoT dataset, for cold-start supervised fine-tuning. Following this initialization, we introduced a FAS-specific reinforcement learning paradigm. Within this stage, Degradation-Simulated Augmentation (DSA) places paired clean and synthetically degraded trajectories into the same rollout group, forcing the model to anchor on stable spoofing evidence rather than incidental image variations. Furthermore, Difficulty-Aware GRPO (DA-GRPO) dynamically tracks the proficiency of semantic task-attack subgroups and adaptively reweights advantages for unreliable categories, preventing easy-sample dominance and ensuring complex attacks are fully optimized. Extensive experiments demonstrate this two-stage approach achieves state-of-the-art multi-task accuracy, cross-domain generalization, and rationale quality.

\bibliography{aaai2027}

@inproceedings{Maatta2011MicroTexture,
  title     = {Face Spoofing Detection from Single Images Using Micro-Texture Analysis},
  author    = {M{\"a}{\"a}tt{\"a}, Jukka and Hadid, Abdenour and Pietik{\"a}inen, Matti},
  booktitle = {International Joint Conference on Biometrics (IJCB)},
  year      = {2011}
}

@inproceedings{Chingovska2012LBP,
  title     = {On the Effectiveness of Local Binary Patterns in Face Anti-Spoofing},
  author    = {Chingovska, Ivana and Anjos, Andr{\'e} and Marcel, S{\'e}bastien},
  booktitle = {BIOSIG},
  year      = {2012}
}

@inproceedings{Yu2020CDCN,
  title     = {Searching Central Difference Convolutional Networks for Face Anti-Spoofing},
  author    = {Yu, Zitong and Zhao, Chenxu and Wang, Zezheng and Qin, Yunxiao and Su, Zhuo and Li, Xiaobai and Zhou, Feng and Zhao, Guoying},
  booktitle = {Proceedings of the IEEE/CVF Conference on Computer Vision and Pattern Recognition (CVPR)},
  year      = {2020}
}

@inproceedings{Liu2023UDGFAS,
  title     = {Towards Unsupervised Domain Generalization for Face Anti-Spoofing},
  author    = {Liu, Y. and others},
  booktitle = {Proceedings of the IEEE/CVF International Conference on Computer Vision (ICCV)},
  year      = {2023}
}

@inproceedings{Radford2021CLIP,
  title     = {Learning Transferable Visual Models From Natural Language Supervision},
  author    = {Radford, Alec and Kim, Jong Wook and Hallacy, Chris and others},
  booktitle = {Proceedings of the 38th International Conference on Machine Learning (ICML)},
  year      = {2021}
}

@inproceedings{Srivatsan2023FLIP,
  title     = {FLIP: Cross-domain Face Anti-spoofing with Language Guidance},
  author    = {Srivatsan, Koushik and Naseer, Muzammal and Nandakumar, Karthik},
  booktitle = {Proceedings of the IEEE/CVF International Conference on Computer Vision (ICCV)},
  year      = {2023}
}

@inproceedings{Liu2024CFPLFAS,
  title     = {CFPL-FAS: Class Free Prompt Learning for Generalizable Face Anti-spoofing},
  author    = {Liu, Ajian and Xue, Shuai and Gan, Jianwen and others},
  booktitle = {Proceedings of the IEEE/CVF Conference on Computer Vision and Pattern Recognition (CVPR)},
  year      = {2024}
}

@inproceedings{Liu2024BUDoPT,
  title     = {Bottom-Up Domain Prompt Tuning for Generalized Face Anti-Spoofing},
  author    = {Liu, Si-Qi and Wang, Qirui and Yuen, Pong C.},
  booktitle = {European Conference on Computer Vision (ECCV)},
  year      = {2024}
}

@article{Mu2023TeGDG,
  title   = {TeG-DG: Textually Guided Domain Generalization for Face Anti-Spoofing},
  author  = {Mu, Lianrui and Bai, Jianhong and He, Xiaoxuan and others},
  journal = {arXiv preprint arXiv:2311.18420},
  year    = {2023}
}

@inproceedings{Liu2023LLaVA,
  title     = {Visual Instruction Tuning},
  author    = {Liu, Haotian and Li, Chunyuan and Wu, Qingyang and Lee, Yong Jae},
  booktitle = {Advances in Neural Information Processing Systems (NeurIPS)},
  year      = {2023}
}

@article{shao2024deepseekmath,
  title         = {DeepSeekMath: Pushing the Limits of Mathematical Reasoning in Open Language Models},
  author        = {Shao, Zhihong and Wang, Peiyi and Zhu, Qihao and Xu, Runxin and Song, Junxiao and Bi, Xiao and Zhang, Haowei and Zhang, Mingchuan and Li, Y.K. and Wu, Y. and Guo, Daya},
  journal       = {arXiv preprint arXiv:2402.03300},
  year          = {2024},
  eprint        = {2402.03300},
  archivePrefix = {arXiv},
  primaryClass  = {cs.CL},
  url           = {https://arxiv.org/abs/2402.03300}
}

@article{zheng2024llamafactory,
  title         = {LlamaFactory: Unified Efficient Fine-Tuning of 100+ Language Models},
  author        = {Zheng, Yaowei and Zhang, Richong and Zhang, Jun and Ye, Zheng and Luo, Ning and Jin, Yucheng and others},
  journal       = {arXiv preprint arXiv:2403.13372},
  year          = {2024},
  eprint        = {2403.13372},
  archivePrefix = {arXiv},
  primaryClass  = {cs.CL}
}

@article{bai2025qwen25vl,
  title         = {Qwen2.5-VL Technical Report},
  author        = {Bai, Shuai and Chen, Keqin and Liu, Xuejing and Wang, Jialin and Ge, Wenbin and Song, Sibo and Dang, Kai and Wang, Peng and Wang, Shijie and Tang, Jun and Zhong, Humen and Zhu, Yuanzhi and Yang, Mingkun and Li, Zhaohai and Wan, Jianqiang and Wang, Pengfei and Ding, Wei and Fu, Zheren and Xu, Yiheng and Ye, Jiabo and Zhang, Xi and Xie, Tianbao and Cheng, Zesen and Zhang, Hang and Yang, Zhibo and Xu, Haiyang and Lin, Junyang and others},
  journal       = {arXiv preprint arXiv:2502.13923},
  year          = {2025},
  eprint        = {2502.13923},
  archivePrefix = {arXiv},
  primaryClass  = {cs.CV}
}

@article{wang2025faceshield,
  title         = {FaceShield: Explainable Face Anti-Spoofing with Multimodal Large Language Models},
  author        = {Wang, Hongyang and Shi, Yichen and Tao, Zhuofu and Gao, Yuhao and Zhang, Liepiao and Lin, Xun and Feng, Jun and Yuan, Xiaochen and Yu, Zitong and Cao, Xiaochun},
  journal       = {arXiv preprint arXiv:2505.09415},
  year          = {2025},
  eprint        = {2505.09415},
  archivePrefix = {arXiv},
  primaryClass  = {cs.CV}
}

@inproceedings{he2016resnet,
  title     = {Deep Residual Learning for Image Recognition},
  author    = {He, Kaiming and Zhang, Xiangyu and Ren, Shaoqing and Sun, Jian},
  booktitle = {Proceedings of the IEEE Conference on Computer Vision and Pattern Recognition (CVPR)},
  year      = {2016}
}

@inproceedings{wang2022patchnet,
  title     = {PatchNet: A Simple Face Anti-Spoofing Framework via Fine-Grained Patch Recognition},
  author    = {Wang, Chien-Yi and Lu, Yu-Ding and Yang, Shang-Ta and Lai, Shang-Hong},
  booktitle = {Proceedings of the IEEE/CVF Conference on Computer Vision and Pattern Recognition (CVPR)},
  year      = {2022}
}

@inproceedings{zhou2022coop,
  title     = {Conditional Prompt Learning for Vision-Language Models},
  author    = {Zhou, Kaiyang and Yang, Jingkang and Loy, Chen Change and Liu, Ziwei},
  booktitle = {Proceedings of the IEEE/CVF Conference on Computer Vision and Pattern Recognition (CVPR)},
  year      = {2022},
  note      = {CoOp}
}

@article{bai2023qwen-vl,
  title         = {Qwen-VL: A Versatile Vision-Language Model for Understanding, Localization, Text Reading, and Beyond},
  author        = {Bai, Jinze and Bai, Shuai and Yang, Shusheng and Wang, Shijie and Tan, Sinan and Wang, Peng and Lin, Junyang and Zhou, Chang and Zhou, Jingren},
  journal       = {arXiv preprint arXiv:2308.12966},
  year          = {2023}
}

@article{zhu2023minigpt4,
  title         = {MiniGPT-4: Enhancing Vision-Language Understanding with Advanced Large Language Models},
  author        = {Zhu, Deyao and Chen, Jun and Shen, Xiaoqian and Li, Xiang and Elhoseiny, Mohamed},
  journal       = {arXiv preprint arXiv:2304.10592},
  year          = {2023}
}

@article{he2024bunny,
  title         = {Efficient Multimodal Learning from Data-centric Perspective},
  author        = {He, Muyang and Liu, Yexin and Wu, Boya and Yuan, Jianhao and Wang, Yueze and Huang, Tiejun and Zhao, Bo},
  journal       = {arXiv preprint arXiv:2402.11530},
  year          = {2024},
  note          = {Bunny}
}

@misc{anthropic2025claude45,
  title        = {Introducing Claude Sonnet 4.5},
  author       = {{Anthropic}},
  year         = {2025},
  howpublished = {\url{https://www.anthropic.com/news/claude-sonnet-4-5}},
  note         = {Accessed 2026-02-12}
}

@misc{openai2025gpt52systemcard,
  title        = {Update to GPT-5 System Card: GPT-5.2},
  author       = {{OpenAI}},
  year         = {2025},
  howpublished = {\url{https://cdn.openai.com/pdf/3a4153c8-c748-4b71-8e31-aecbde944f8d/oai_5_2_system-card.pdf}},
  note         = {Accessed 2026-02-12}
}

@misc{google2025gemini3pro,
  title        = {Gemini models: Gemini 3 Pro},
  author       = {{Google}},
  year         = {2025},
  howpublished = {\url{https://ai.google.dev/gemini-api/docs/models}},
  note         = {Accessed 2026-02-12}
}

@article{wei2023lenna,
  title         = {Lenna: Language Enhanced Reasoning Detection Assistant},
  author        = {Wei, Fei and Zhang, Xinyu and Zhang, Ailing and Zhang, Bo and Chu, Xiangxiang},
  journal       = {arXiv preprint arXiv:2312.02433},
  year          = {2023},
  eprint        = {2312.02433},
  archivePrefix = {arXiv},
  primaryClass  = {cs.CV}
}

@inproceedings{zhou2023iadg,
  title     = {Instance-Aware Domain Generalization for Face Anti-Spoofing},
  author    = {Zhou, Qianyu and Zhang, Ke-Yue and Yao, Taiping and Lu, Xuequan and Yi, Ran and Ding, Shouhong and Ma, Lizhuang},
  booktitle = {Proceedings of the IEEE/CVF Conference on Computer Vision and Pattern Recognition (CVPR)},
  year      = {2023}
}

@article{cai2024fasaug,
  title         = {Towards Data-Centric Face Anti-Spoofing: Improving Cross-domain Generalization via Physics-based Data Synthesis},
  author        = {Cai, Rizhao and Soh, Cecelia and Yu, Zitong and Li, Haoliang and Yang, Wenhan and Kot, Alex},
  journal       = {arXiv preprint arXiv:2409.03501},
  year          = {2024}
}

@article{lin2023sphinx,
  title={Sphinx: The joint mixing of weights, tasks, and visual embeddings for multi-modal large language models},
  author={Lin, Ziyi and Liu, Chris and Zhang, Renrui and Gao, Peng and Qiu, Longtian and Xiao, Han and Qiu, Han and Lin, Chen and Shao, Wenqi and Chen, Keqin and others},
  journal={arXiv preprint arXiv:2311.07575},
  year={2023}
}

@inproceedings{tan2026veritas,
  title     = {Veritas: Generalizable Deepfake Detection via Pattern-Aware Reasoning},
  author    = {Tan, Hao and Lan, Jun and Tan, Zichang and Shi, Senyuan and Liu, Ajian and Song, Chuanbiao and Zhu, Huijia and Wang, Weiqiang and Wan, Jun and Lei, Zhen},
  booktitle = {International Conference on Learning Representations (ICLR)},
  year      = {2026},
  note      = {Oral},
  url       = {https://openreview.net/forum?id=5VXJPS1HoM}
}

@inproceedings{zhang2025interpretable,
  title={Interpretable face anti-spoofing: Enhancing generalization with multimodal large language models},
  author={Zhang, Guosheng and Wang, Keyao and Yue, Haixiao and Liu, Ajian and Zhang, Gang and Yao, Kun and Ding, Errui and Wang, Jingdong},
  booktitle={Proceedings of the AAAI Conference on Artificial Intelligence},
  volume={39},
  pages={9896--9904},
  year={2025}
}

@misc{zhang2026harnessingcotfas,
  title         = {Harnessing Chain-of-Thought Reasoning in Multimodal Large Language Models for Face Anti-Spoofing},
  author        = {Honglu Zhang and Zhiqin Fang and Ningning Zhao and Saihui Hou and Long Ma and Renwang Pei and Zhaofeng He},
  year          = {2026},
  eprint        = {2506.01783},
  archivePrefix = {arXiv},
  primaryClass  = {cs.CV},
  note          = {Accepted to CVPR 2026},
  doi           = {10.48550/arXiv.2506.01783}
}

@misc{jiang2025tasksolving,
  title         = {Exploring Task-Solving Paradigm for Generalized Cross-Domain Face Anti-Spoofing via Reinforcement Fine-Tuning},
  author        = {Fangling Jiang and Qi Li and Weining Wang and Gang Wang and Bing Liu and Zhenan Sun},
  year          = {2025},
  eprint        = {2506.21895},
  archivePrefix = {arXiv},
  primaryClass  = {cs.CV},
  doi           = {10.48550/arXiv.2506.21895}
}

@article{yu2022deep,
  title={Deep learning for face anti-spoofing: A survey},
  author={Yu, Zitong and Qin, Yunxiao and Li, Xiaobai and Zhao, Chenxu and Lei, Zhen and Zhao, Guoying},
  journal={IEEE transactions on pattern analysis and machine intelligence},
  volume={45},
  number={5},
  pages={5609--5631},
  year={2022},
  publisher={IEEE}
}

@inproceedings{AtoumIJCB2017,
  title     = {Face Anti-Spoofing Using Patch and Depth-Based CNNs},
  author    = {Atoum, Yousef and Liu, Yaojie and Jourabloo, Amin and Liu, Xiaoming},
  booktitle = {International Joint Conference on Biometrics (IJCB)},
  pages     = {319--328},
  year      = {2017},
  doi       = {10.1109/BTAS.2017.8272713}
}

@inproceedings{GeorgeICB2019,
  title     = {Deep Pixel-wise Binary Supervision for Face Presentation Attack Detection},
  author    = {George, Anjith and Marcel, S{\'e}bastien},
  booktitle = {International Conference on Biometrics (ICB)},
  year      = {2019},
  doi       = {10.1109/ICB45273.2019.8987370}
}

@inproceedings{YangCVPR2019,
  title     = {Face Anti-Spoofing: Model Matters, so Does Data},
  author    = {Yang, Xiao and Luo, Wenhan and Bao, Linchao and Gao, Yuan and Gong, Dihong and Zheng, Shibao and Li, Zhifeng and Liu, Wei},
  booktitle = {Proceedings of the IEEE/CVF Conference on Computer Vision and Pattern Recognition (CVPR)},
  pages     = {3507--3516},
  year      = {2019},
  doi       = {10.1109/CVPR.2019.00362}
}

@inproceedings{LiuCVPR2019DeepTree,
  title     = {Deep Tree Learning for Zero-Shot Face Anti-Spoofing},
  author    = {Liu, Yaojie and Stehouwer, Joel and Jourabloo, Amin and Liu, Xiaoming},
  booktitle = {Proceedings of the IEEE/CVF Conference on Computer Vision and Pattern Recognition (CVPR)},
  year      = {2019},
  doi       = {10.1109/CVPR.2019.00481}
}

@article{george2019wmca,
  title   = {Biometric Face Presentation Attack Detection with Multi-Channel Convolutional Neural Network},
  author  = {George, Anjith and Mostaani, Zohreh and Geissenbuhler, David and Nikisins, Olegs and Anjos, Andr{\'e} and Marcel, S{\'e}bastien},
  journal = {IEEE Transactions on Information Forensics and Security},
  volume  = {15},
  pages   = {42--55},
  year    = {2019}
}

@inproceedings{rostami2021padisi,
  title     = {Detection and Continual Learning of Novel Face Presentation Attacks},
  author    = {Rostami, Mohammad and Spinoulas, Leonidas and Hussein, Mohamed and Mathai, Joe and Abd-Almageed, Wael},
  booktitle = {Proceedings of the IEEE/CVF International Conference on Computer Vision (ICCV)},
  year      = {2021}
}

@article{guo2022siwmv2,
  title   = {Multi-domain Learning for Updating Face Anti-spoofing Models},
  author  = {Guo, Xiao and Liu, Yaojie and Jain, Anil and Liu, Xiaoming},
  journal = {arXiv preprint arXiv:2208.11148},
  year    = {2022},
  doi     = {10.48550/arXiv.2208.11148}
}

@misc{gemini25flash,
  title        = {Gemini 2.5 Flash},
  author       = {{Google}},
  year         = {2025},
  howpublished = {\url{https://ai.google.dev/}},
  note         = {Accessed: 2026-03-03}
}

@misc{gpt5,
  title        = {GPT-5},
  author       = {{OpenAI}},
  year         = {2025},
  howpublished = {\url{https://openai.com/}},
  note         = {Accessed: 2026-03-03}
}

@article{shi2025shield,
  title   = {SHIELD: An Evaluation Benchmark for Face Spoofing and Forgery Detection with Multimodal Large Language Models},
  author  = {Shi, Yichen and Gao, Yuhao and Lai, Yingxin and Wang, Hongyang and Feng, Jun and He, Lei and Wan, Jun and Chen, Changsheng and Yu, Zitong and Cao, Xiaochun},
  journal = {Visual Intelligence},
  year    = {2025},
  eprint  = {2402.04178},
  archivePrefix = {arXiv},
  primaryClass = {cs.CV},
  doi = {10.48550/arXiv.2402.04178}
}

@misc{zhang2026tarfash,
  title         = {From Intuition to Investigation: A Tool-Augmented Reasoning MLLM Framework for Generalizable Face Anti-Spoofing},
  author        = {Zhang, Haoyuan and Wang, Keyao and Zhang, Guosheng and Yue, Haixiao and Tan, Zhiwen and Peng, Siran and Zhang, Tianshuo and Tan, Xiao and Chen, Kunbin and He, Wei and Wang, Jingdong and Liu, Ajian and Zhu, Xiangyu and Lei, Zhen},
  year          = {2026},
  eprint        = {2603.01038},
  archivePrefix = {arXiv},
  primaryClass  = {cs.CV},
  note          = {arXiv preprint}
}

@inproceedings{wang2025fsfm,
  title     = {FSFM: A Generalizable Face Security Foundation Model via Self-Supervised Facial Representation Learning},
  author    = {Wang, Gaojian and Lin, Feng and Wu, Tong and Liu, Zhenguang and Ba, Zhongjie and Ren, Kui},
  booktitle = {Proceedings of the IEEE/CVF Conference on Computer Vision and Pattern Recognition (CVPR)},
  year      = {2025},
  doi       = {10.1109/CVPR52734.2025.02269}
}

@inproceedings{le2024gacfas,
  title     = {Gradient Alignment for Cross-Domain Face Anti-Spoofing},
  author    = {Le, Binh M. and Woo, Simon S.},
  booktitle = {Proceedings of the IEEE/CVF Conference on Computer Vision and Pattern Recognition (CVPR)},
  year      = {2024},
  eprint    = {2402.18817},
  archivePrefix = {arXiv},
  primaryClass = {cs.CV},
  doi       = {10.48550/arXiv.2402.18817}
}

@inproceedings{ma2026pafas,
  title     = {PA-FAS: Towards Interpretable and Generalizable Multimodal Face Anti-Spoofing via Path-Augmented Reinforcement Learning},
  author    = {Ma, Yingjie and Lin, Xun and Xu, Yong and Xie, Weicheng and Yu, Zitong},
  booktitle = {Proceedings of the AAAI Conference on Artificial Intelligence},
  year      = {2026},
  note      = {Oral},
  eprint    = {2511.17927},
  archivePrefix = {arXiv},
  primaryClass = {cs.CV},
  doi       = {10.48550/arXiv.2511.17927}
}

@inproceedings{yu2025mvpfas,
  title     = {Multi-View Slot Attention Using Paraphrased Texts for Face Anti-Spoofing},
  author    = {Yu, Jeongmin and Kim, Susang and Lee, Kisu and Kwon, Taekyoung and Shin, Won-Yong and Kim, Ha Young},
  booktitle = {Proceedings of the IEEE/CVF International Conference on Computer Vision},
  pages     = {21117--21128},
  year      = {2025}
}

@article{yu2025dapo,
  title        = {{DAPO}: An Open-Source {LLM} Reinforcement Learning System at Scale},
  author       = {Yu, Qiying and Zhang, Zheng and Zhu, Ruofei and Yuan, Yufeng and Zuo, Xiaochen and Yue, Yu and Dai, Weinan and Fan, Tiantian and Liu, Gaohong and Liu, Lingjun and Liu, Xin and Lin, Haibin and Lin, Zhiqi and Ma, Bole and Sheng, Guangming and Tong, Yuxuan and Zhang, Chi and Zhang, Mofan and Zhang, Wang and Zhu, Hang and Zhu, Jinhua and Chen, Jiaze and Chen, Jiangjie and Wang, Chengyi and Yu, Hongli and Song, Yuxuan and Wei, Xiangpeng and Zhou, Hao and Liu, Jingjing and Ma, Wei-Ying and Zhang, Ya-Qin and Yan, Lin and Qiao, Mu and Wu, Yonghui and Wang, Mingxuan},
  journal      = {arXiv preprint arXiv:2503.14476},
  year         = {2025}
}

@article{zheng2025gspo,
  title        = {Group Sequence Policy Optimization},
  author       = {Zheng, Chujie and Liu, Shixuan and Li, Mingze and Chen, Xiong-Hui and Yu, Bowen and Gao, Chang and Dang, Kai and Liu, Yuqiong and Men, Rui and Yang, An and Zhou, Jingren and Lin, Junyang},
  journal      = {arXiv preprint arXiv:2507.18071},
  year         = {2025}
}

\end{document}